%% file: main.tex
\documentclass{article} 
\usepackage{iclr2026_conference,times}

\input{math_commands.tex}

\usepackage{hyperref}
\usepackage{url}

\usepackage{graphicx}
\usepackage{amsmath,amssymb}
\usepackage{booktabs}
\usepackage{float}

\title{Pretraining Large Language Models with MXFP4 on Native FP4 Hardware}

\author{
Musa Cim\textsuperscript{1} \quad
Sarthak Arora\textsuperscript{2} \quad
Poovaiah Palangappa\textsuperscript{2} \quad
Miro Hodak\textsuperscript{2} \\
\textbf{Ravi Dwivedula}\textsuperscript{2} \quad
\textbf{Meena Arunachalam}\textsuperscript{2} \quad
\textbf{Mahmut Taylan Kandemir}\textsuperscript{1} \\[0.5em]
\textsuperscript{1}The Pennsylvania State University \quad
\textsuperscript{2}Advanced Micro Devices, Inc. \\[0.3em]
\texttt{\{mtc5693, mtk2\}@psu.edu} \\
\texttt{\{Sarthak.Arora, Poovaiah.Palangappa, Miro.Hodak,} \\
\texttt{Ravi.Dwivedula, Meena.Arunachalam\}@amd.com}
}

\iclrfinalcopy 
\begin{document}

\maketitle

\begin{abstract}
Why does full-pipeline FP4 training of large language models often diverge, even when forward activations and activation gradients remain stable? We address this question through a controlled study of MXFP4 quantization in transformer training, progressively enabling FP4 across forward propagation (Fprop), activation gradients (Dgrad), and weight gradients (Wgrad) while holding all other factors fixed. In full pretraining of Llama 3.1–8B on the C4 dataset, we observe that quantizing Wgrad is the primary driver of convergence degradation, whereas FP4 in Fprop and Dgrad alone introduces only modest additional token requirements. To interpret this behavior, we evaluate both structured and stochastic interventions under a controlled experimental setting. We find that stochastic rounding and randomized Hadamard rotations fail to stabilize training once Wgrad is quantized, whereas deterministic Hadamard rotations consistently restore stable optimization. These results suggest that FP4 training instability is driven by structured micro-scaling errors along sensitive gradient paths, rather than by insufficient stochasticity. We run experiments with native MXFP4 support on AMD Instinct MI355X GPUs, enabling controlled investigation of these effects without reliance on software emulation.
\end{abstract}

\section{Introduction}
The rapid growth of large language models (LLMs), including Llama~3~\citep{grattafiori2024llama}, has intensified the need for more efficient, hardware-aware training algorithms. While standard half-precision formats impose substantial demands on memory bandwidth and computational resources, FP8 has recently emerged as a practical low-precision alternative. Nevertheless, training with 4-bit formats, e.g.,  microscaling formats, NVFP4, and MXFP4, remains a significant challenge.
A key difficulty in 4-bit training is that activation and gradient outliers amplify the quantization error and can destabilize optimization. When activation outliers are coarsely quantized, they can dominate the quantization range and severely degrade low-bit floating-point quantization quality, as observed in early FP4 studies~\citep{liu2023llm}. \citet{zhou2025towards} has explored FP4 pretraining recipes via module-wise mixed precision and staged schedules to mitigate FP4 quantization noise, and \citet{abecassis2025pretraining} has investigated pretraining with vendor-specific formats such as NVFP4.

In this work, we ask whether we can replace FP8 GEMM kernels in transformer linear layers with MXFP4 GEMM kernels while preserving training quality, and we use controlled experiments to identify which training components (Fprop, Dgrad, Wgrad) most strongly determine stability and what stabilizers are effective. To answer this question, we progressively enable MXFP4 across Fprop, Dgrad, and Wgrad (holding all other factors fixed) and use this controlled ladder to validate or falsify candidate stabilizers.

Our primary motivation is {\em efficiency}: with native FP4 tensor-core support, MXFP4 GEMMs can provide higher throughput and reduce memory bandwidth pressure compared to FP8. However, replacing FP8 with MXFP4 can increase token-to-converge or even destabilize training. Beyond efficiency, understanding where low-precision training fails and why provides actionable insight into the optimization dynamics of large transformers under extreme quantization. Here, we use controlled, stage-wise MXFP4 enablement to identify where FP4 training breaks, and we use Hadamard rotations as a structured probe to interpret which error sources drive instability. We report the following findings:

\begin{itemize}
\item \textbf{Where:} Progressive MXFP4 enablement reveals that convergence instability is dominated by Wgrad quantization, whereas MXFP4 in Fprop and Dgrad alone remains comparatively stable.
\item \textbf{Why:} Stochastic interventions fail to stabilize training when Wgrad is quantized, suggesting that added randomness amplifies the effective quantization error rather than mitigating outlier effects.
\item \textbf{How:} Deterministic Hadamard rotations restore stable optimization in the full MXFP4 pipeline under the same controlled setting where stochastic variants fail, making them a practical probe for isolating stabilizing structure in FP4 training.
\end{itemize}

\section{Experimental Setup}
\subsection{Native FP4 Hardware Support}
Prior work on FP4~\citep{chmiel2025fp4, cim2026diagnosing, wang2025optimizing} has often relied on software emulation, which introduces additional overhead and can obscure both performance characteristics and training stability. Our experiments instead use native FP4 tensor support on AMD Instinct MI355X GPUs, enabling MXFP4 computation without emulation. To our knowledge, this is the first demonstration of full MXFP4 pretraining on the MI355X-class hardware~\citep{amd_mi355x_specs}.

\subsection{Models and Tasks}
We follow the MLPerf LLM pretraining setup on C4 dataset and train Llama~3.1--8B. We define convergence as reaching validation perplexity 3.3. For each method, we report the number of training tokens required to first reach perplexity $\leq 3.3$. We report token overhead as the relative increase in tokens compared to the FP8 baseline at this threshold.

\section{Evaluation Results}
\subsection{Hypotheses}
Our experiments test four hypotheses about full MXFP4 pretraining stability: {\bf (H1)} progressive quantization reveals that degrading convergence is dominated by quantizing Wgrad relative to Fprop or Dgrad; {\bf (H2)} stochastic methods (stochastic rounding or randomized Hadamard) provide limited benefit because they add noise without addressing the underlying outlier-driven error in MXFP4 micro-scaling; {\bf (H3)} a deterministic Hadamard rotation is consistent with reducing the impact of outliers entering MXFP4 blocks and thereby stabilizes optimization in the full pipeline; and {\bf (H4)} enabling MXFP4 in more GEMM paths (stage-wise MXFP4 enablement) increases training-step throughput, and when token overhead remains small, this translates to an end-to-end speedup at the MLPerf target.

\subsection{Stage-Wise MXFP4 enablement}
To isolate which components of MXFP4 training drive convergence degradation, we progressively enable MXFP4 in the training pipeline and measure token overhead to reach validation perplexity 3.3 (Table~\ref{tab:ladder_expanded_rowwise}). Relative to FP8, applying MXFP4 to Fprop only incurs an 8--9\% token overhead, and extending MXFP4 to Fprop + Dgrad yields a similar 10--11\% overhead. In contrast, enabling MXFP4 for Fprop + Dgrad + Wgrad increases the overhead to 26--27\%, indicating that the Wgrad quantization is the dominant contributor to convergence degradation. We then evaluate stabilization strategies under the same progressive enabling. Stochastic rounding matches the unstabilized overhead for Fprop and Fprop+Dgrad, but does not converge when applied to the full pipeline (Fprop+Dgrad+Wgrad). Randomized Hadamard similarly matches the partial-stage overhead but does not converge for the full pipeline. In contrast, applying a deterministic Hadamard transform to the full MXFP4 pipeline restores stable training and reduces the token overhead to 8--9\% while maintaining the same target perplexity; we additionally find $H_{16}$ (a $16 \times 16$ Hadamard matrix) is 8\% faster than $H_{32}$ ($32 \times 32$) in our kernels (1.08$\times$ vs 1.00$\times$).

\begin{table*}[t]
   \caption{Stage-wise MXFP4 enablement with stabilization strategies. Token overhead is relative to the FP8 baseline (validation perplexity 3.3). ``Does Not Converge'' indicates the run diverged or failed to reach the perplexity-3.3 target despite extended training.}
   \label{tab:ladder_expanded_rowwise}
   \centering
   \begin{tabular}{lllc}
   \toprule
   \textbf{Stabilizer} & \textbf{Hadamard} & \textbf{MXFP4 GEMMs (others FP8)} & \textbf{Token overhead} \\
   \midrule
   \textbf{FP8 Baseline} & -- & None (All GEMMs FP8) & 0\% \\
   \midrule
   
   None & -- & Fprop & 8--9\% \\
   None & -- & Fprop + Dgrad & 10--11\% \\
   None & -- & Fprop + Dgrad + Wgrad & 26--27\% \\
   \midrule
   
   Stochastic Rounding & -- & Fprop & 8--9\% \\
   Stochastic Rounding & -- & Fprop + Dgrad & 10--11\% \\
   Stochastic Rounding & -- & Fprop + Dgrad + Wgrad & Does Not Converge \\
   \midrule
   
   Randomized Hadamard & $H_{16}$ & Fprop & 8--9\% \\
   Randomized Hadamard & $H_{16}$ & Fprop + Dgrad & 10--11\% \\
   Randomized Hadamard & $H_{16}$ & Fprop + Dgrad + Wgrad & Does Not Converge \\
   \midrule
   
   Deterministic Hadamard & $H_{32}$ & Fprop & 8--9\% \\
   Deterministic Hadamard & $H_{32}$ & Fprop + Dgrad & 10--11\% \\
   Deterministic Hadamard & $H_{32}$ & Fprop + Dgrad + Wgrad & 8--9\% \\
   \midrule
   
   Deterministic Hadamard & $H_{16}$ & Fprop & 8--9\% \\
   Deterministic Hadamard & $H_{16}$ & Fprop + Dgrad & 10--11\% \\
   \textbf{Deterministic Hadamard} & \textbf{$H_{16}$} & \textbf{Fprop + Dgrad + Wgrad} & \textbf{8--9\%} \\
   \bottomrule
   \end{tabular}
   
   \vspace{0.5em}
   \small
   \textbf{Kernel throughput:} $H_{16}$ is 8\% faster than $H_{32}$ (1.08$\times$ vs 1.00$\times$).
   \end{table*}

\begin{figure}[H]
\centering
\begin{minipage}[t]{0.48\linewidth}
  \vspace{0pt}
  \centering
  \includegraphics[width=\linewidth]{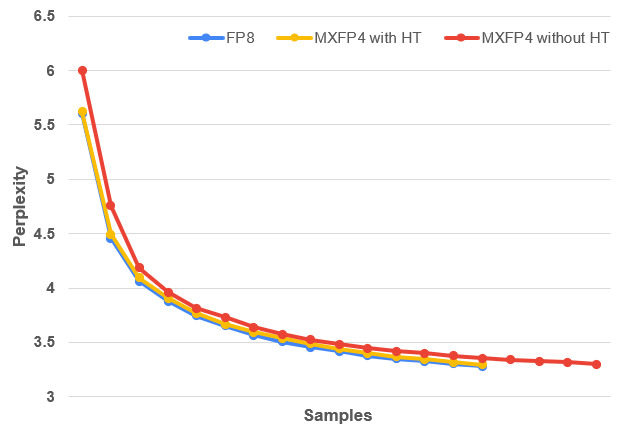}
  \caption{Validation perplexity vs.\ training tokens for Llama~3.1--8B under MLPerf pretraining on C4 dataset. We compare FP8, full-pipeline MXFP4 (Fprop + Dgrad + Wgrad; no stabilizer), and full-pipeline MXFP4 + deterministic Hadamard ($H_{16}$). MXFP4 + deterministic Hadamard closely tracks FP8, while full-pipeline MXFP4 without stabilization converges more slowly and is less stable.}
  \label{fig:convergence}
\end{minipage}\hfill
\begin{minipage}[t]{0.48\linewidth}
  \vspace{0pt}
  \centering
  \includegraphics[width=\linewidth]{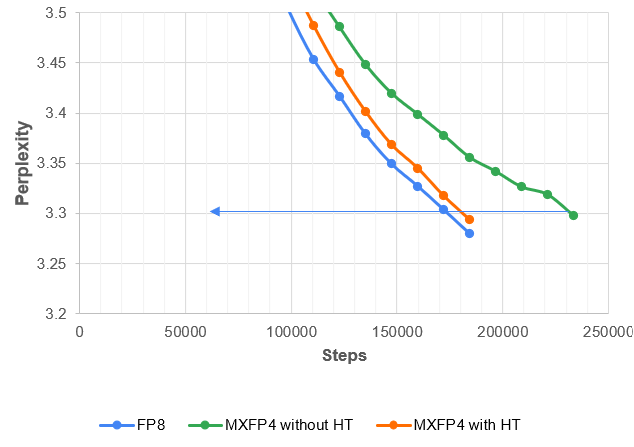}
  \caption{Zoomed-in view of later training (from Figure~\ref{fig:convergence}). The MLPerf target is perplexity 3.3. Deterministic Hadamard ($H_{16}$) maintains tight alignment with the FP8 baseline compared to the unstabilized MXFP4 run.}
  \label{fig:convergence_zoom}
\end{minipage}
\end{figure}

\subsection{Convergence Analysis}
We now visualize the training trajectories underlying the token-overhead results in Table~\ref{tab:ladder_expanded_rowwise}. Specifically, we plot validation perplexity as a function of training tokens and evaluate whether each configuration reaches the MLPerf target (perplexity 3.3) in a stable fashion. Consistent with stage-wise enablement, naive MXFP4 degrades convergence most strongly once Wgrad is quantized, while deterministic Hadamard stabilizes the full MXFP4 pipeline and yields trajectories that closely track the FP8 baseline.

\begin{center}
   \fcolorbox{blue}{white}{
     \parbox{0.94\linewidth}{
      \textbf{Key Understanding \#1:} Progressive MXFP4 enabling shows that convergence degradation is dominated by quantizing Wgrad. Deterministic Hadamard enables stable full-pipeline MXFP4 training; stochastic techniques exhibit higher quantization error when Wgrad is quantized.
     }
   }
   \end{center} 

Additionally, we examine whether restoring convergence stability allows the original efficiency motivation of FP4 to be realized. For the stable configuration (MXFP4 with deterministic Hadamard applied to Fprop, Dgrad, and Wgrad), we observe higher training-step throughput than the FP8 baseline under the same MLPerf C4 dataset setup on MI355X GPUs with native FP4 support (Table~2). Importantly, this throughput advantage translates into an end-to-end speedup only once the dominant instability from Wgrad quantization is controlled; otherwise, increased token-to-target or divergence offsets the raw compute benefit. This observation highlights that the practical speedups for FP4 are stability-gated rather than automatic. 

\begin{table}[H]
\caption{End-to-end training efficiency (MXFP4 + $H_{16}$ vs.\ FP8 baseline).}
\label{tab:throughput}
\centering
\begin{tabular}{lc}
\toprule
\textbf{Metric} & \textbf{MXFP4 + $H_{16}$ vs.\ FP8} \\
\midrule
Train step throughput & +20\% \\
Token overhead to converge & +8--9\% \\
\textbf{End-to-end speedup} & \textbf{+9--10\%} \\
\bottomrule
\end{tabular}
\end{table}

\begin{center}
\fcolorbox{blue}{white}{
\parbox{0.94\linewidth}{
\textbf{Key Understanding \#2:} FP4's practical speedups are stability-gated: without stabilizing Wgrad, increased token-to-target cancels the step-throughput benefit, masking the underlying hardware efficiency of FP4.
}}
\end{center}

Both FP8 and MXFP4+$H_{16}$ runs employ the same MLPerf C4 dataset pretraining setup; values are normalized by the FP8 end-to-end tokens/s measured under identical node, batch, and training configuration. Reduced activation/weight size lowers bandwidth pressure, which is a key bottleneck in large-batch training, and our method additionally stabilizes training in the full pipeline (Table~\ref{tab:ladder_expanded_rowwise}).

\section{Discussion and Future Work}
Our results suggest that Wgrad is the most sensitive component in full-pipeline MXFP4 training, and that stabilizing FP4 pretraining requires controlling micro-scaling error rather than injecting additional stochasticity. At the same time, the recipe is not universal: The FP4 training behavior can vary between different models, datasets, and adaptation methods.

\begin{center}
\fcolorbox{blue}{white}{
   \parbox{0.94\linewidth}{
      \textbf{Key Understanding \#3:} MXFP4 training behavior can be highly setting-dependent: a stabilization that works for full pretraining (MLPerf C4 dataset, Llama~3.1--8B) may not generalize to other models or finetuning methods; so, FP4 recipes should be treated as non-universal without further evidence.
   }
}
\end{center}

\newpage
\section{Conclusion}
We presented a controlled study of replacing FP8 GEMM kernels in transformer linear layers with MXFP4, and showed that convergence degradation is dominated by quantizing Wgrad relative to Fprop or Dgrad. For Llama~3.1--8B under the MLPerf C4 dataset setup, a deterministic Hadamard rotation is the only tested intervention that restores stable full-pipeline MXFP4 training while improving end-to-end efficiency. Overall, our results suggest a practical principle for FP4 LLM training: stability hinges on controlling micro-scaling error in the most sensitive gradient paths, not on adding stochasticity.

\bibliography{iclr2026_conference}
\bibliographystyle{iclr2026_conference}

\newpage

\appendix
\renewcommand{\thesection}{Appendix \Alph{section}}
\section{Methodology}
\label{app:methodology}

\subsection{MXFP4 with micro-scaling}
Standard integer quantization typically applies a single scale factor to an entire tensor. In contrast, MXFP4 uses micro-scaling, where a shared exponent is defined for small blocks (e.g., blocks of 32 elements). Let vector $x$ be divided into blocks. For a block $x_B$, the reconstructed value is:
\begin{equation}
x_{MX} = Q_{FP4}\left(\frac{x_B}{2^{E_{\mathrm{shared}}}}\right)\times 2^{E_{\mathrm{shared}}},
\end{equation}
where $E_{\mathrm{shared}}$ is the maximum exponent in the block~\citep{mxfp}, and $Q_{FP4}$ denotes rounding to the nearest representable 4-bit floating-point value.

\subsection{Hadamard transform architecture}
We use a deterministic Hadamard transform to improve convergence stability and reduce the impact of outliers~\citep{hadamard}. Although randomized Hadamard rotations (e.g., random sign flips) have been widely used, our experiments found that randomized signs can adversely affect convergence stability in this setting. We therefore use a deterministic transform.

A Hadamard matrix $H$ is an orthogonal matrix with entries $\pm 1$. Applying $H$ can be interpreted as a rotation in feature space that disperses concentrated energy across more dimensions. In our implementation we use a tiled layout and choose a Hadamard transform size of 16 ($H_{16}$). We found $H_{16}$ to provide higher kernel throughput on our target architecture, and we find it empirically maintains stable convergence.

\begin{figure}[h]
\centering
\includegraphics[width=0.8\linewidth]{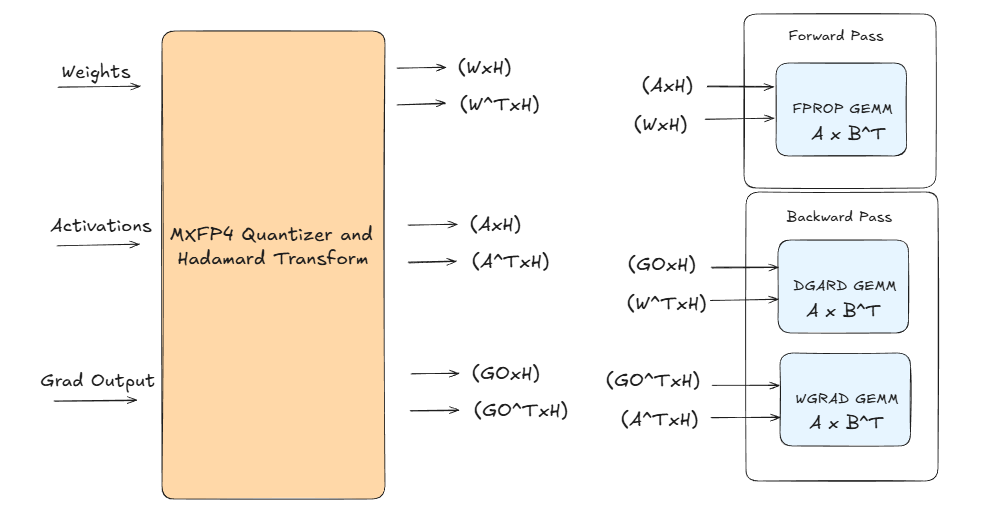}
\caption{Hadamard-transformed MXFP4 architecture for forward and backward passes. Inputs to GEMM kernels are rotated by $H$, and the rotation is reversed during matrix multiplication via $HH^T = I$.}
\label{fig:arch}
\end{figure}

Our pipeline injects Hadamard rotations before GEMM (see Figure~\ref{fig:arch}). Since the transform is linear and orthogonal, it modifies the distribution of data entering the quantizer without altering the underlying computation.
Concretely, we replace the FP8 GEMM kernels used in transformer linear layers with MXFP4 GEMM kernels (for selected passes: Fprop, Dgrad, and/or Wgrad) using AMD's ROCm Transformer Engine; these kernels are exercised through standard transformer linear modules in our training stack.

\section{Architecture Diagram}

\section{Mathematical Proof of Cancellation}
\label{app:cancellation}
To ensure that the Hadamard transformation does not change the arithmetic of linear layers, we use the orthogonality property:
\begin{equation}
HH^T = H^T H = I,
\end{equation}
where $I$ is the identity matrix.

\subsection{Forward pass (FPROP)}
In a standard linear layer, $Y = XW^T$. We apply the transform to both inputs: $\tilde{X} = XH$ and $\tilde{W} = WH$. Then:
\begin{align}
Y_{\mathrm{out}}
&= \tilde{X}(\tilde{W})^T \nonumber\\
&= (XH)(WH)^T \nonumber\\
&= XH(H^T W^T) \nonumber\\
&= X(HH^T)W^T \nonumber\\
&= XW^T.
\end{align}
Thus, Hadamard factors cancel, preserving the exact output while improving quantization robustness.

\subsection{Backward pass (DGRAD)}
For input gradients, the standard operation is $\nabla X = \nabla Y\,W$. With $\tilde{G} = (\nabla Y)H$ and $\tilde{W}_{\mathrm{rot}} = W^T H$:
\begin{align}
\nabla X
&= \tilde{G}(\tilde{W}_{\mathrm{rot}})^T \nonumber\\
&= ((\nabla Y)H)(W^T H)^T \nonumber\\
&= (\nabla Y)H H^T (W^T)^T \nonumber\\
&= (\nabla Y)W.
\end{align}

\subsection{Weight gradient (WGRAD)}
For weight gradients, $\nabla W = (\nabla Y)^T X$. In our pipeline, $\tilde{G}_T = (\nabla Y)^T H$ and $\tilde{X}_T = X^T H$:
\begin{align}
\nabla W
&= \tilde{G}_T(\tilde{X}_T)^T \nonumber\\
&= ((\nabla Y)^T H)(X^T H)^T \nonumber\\
&= (\nabla Y)^T H(H^T X) \nonumber\\
&= (\nabla Y)^T (HH^T)X \nonumber\\
&= (\nabla Y)^T X.
\end{align}
\newpage
\section{1D vs.\ 2D Quantization Strategies}
\label{app:quantization}
We explore 1D and 2D block quantization strategies for activations and gradients:

\textbf{2D block quantization (32$\times$32):} We divide tensors into 32$\times$32 regions and use a per-block \texttt{E8M0} scale for each $C_{\mathrm{in}}\times C_{\mathrm{out}}$ block. This is computationally efficient and transpose-friendly, and enables hardware vectorization while maintaining good precision.

\textbf{1D row-wise quantization (1$\times$32):} This aligns with token-wise processing in transformers. A per-token \texttt{E8M0} scale is computed over 32 consecutive elements, preserving token-level information and working well for activations.

\textbf{1D column-wise quantization (32$\times$1):} This provides fine-grained per-channel scales for weights via a per-$C_{\mathrm{out}}$ scale over 32 elements, but is expensive due to transpose and unfavorable memory access patterns.

\begin{figure}[h]
\centering
\includegraphics[width=0.4\linewidth]{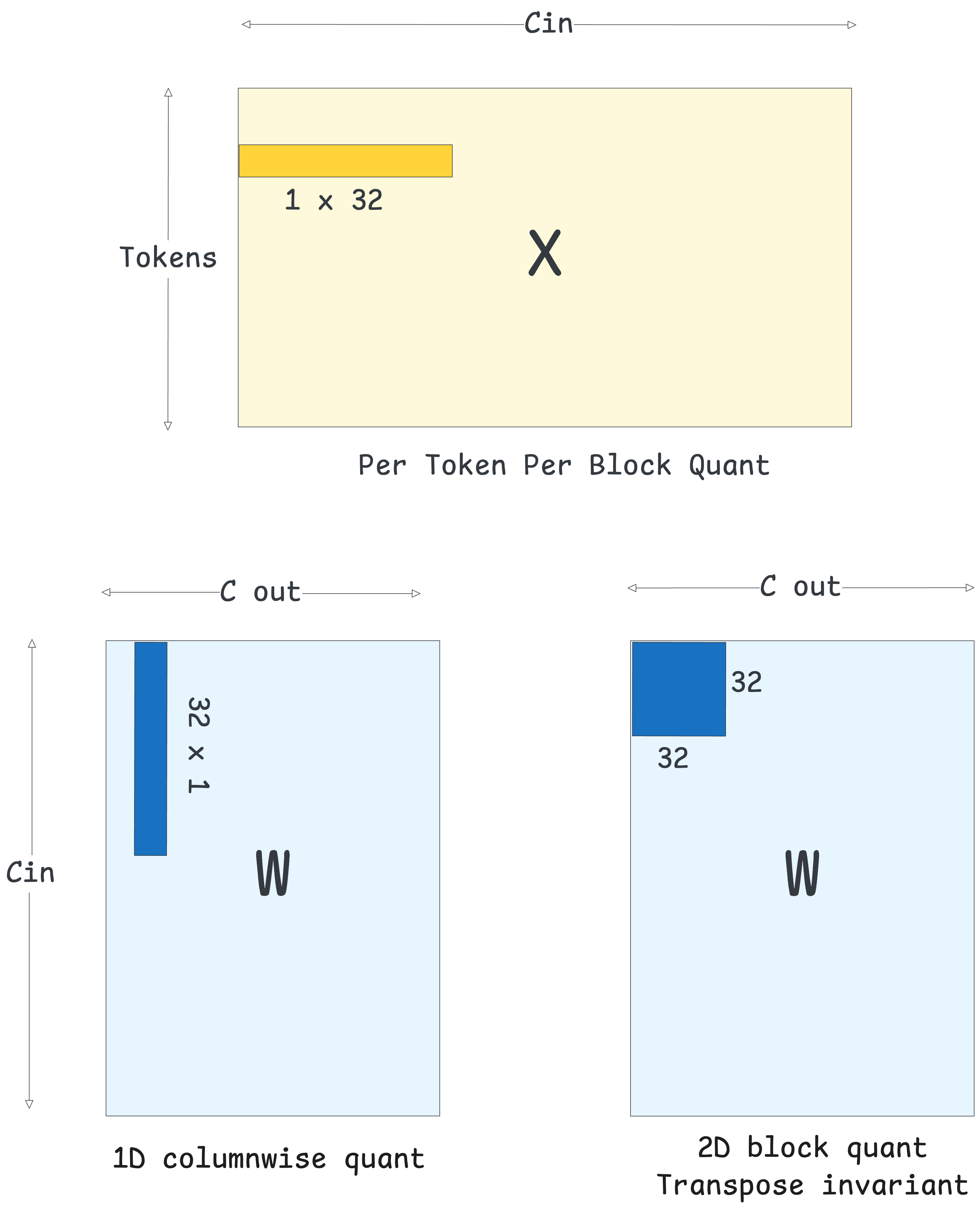}
\caption{Comparison of block quantization strategies: 2D (32$\times$32) is transpose-invariant and efficient; 1D row-wise (1$\times$32) preserves per-token information for activations; 1D column-wise (32$\times$1) provides per-channel scales at higher cost.}
\label{fig:quant_strategies}
\end{figure}

\end{document}

%% file: math_commands.tex
\usepackage{amsmath,amsfonts,bm}

\def\eqref#1{equation~\ref{#1}}

\def\1{\bm{1}}

\DeclareMathAlphabet{\mathsfit}{\encodingdefault}{\sfdefault}{m}{sl}
\SetMathAlphabet{\mathsfit}{bold}{\encodingdefault}{\sfdefault}{bx}{n}



%% file: iclr2026_conference.bib
@article{grattafiori2024llama,
  title={The llama 3 herd of models},
  author={Grattafiori, Aaron and Dubey, Abhimanyu and Jauhri, Abhinav and Pandey, Abhinav and Kadian, Abhishek and Al-Dahle, Ahmad and Letman, Aiesha and Mathur, Akhil and Schelten, Alan and Vaughan, Alex and others},
  journal={arXiv preprint arXiv:2407.21783},
  year={2024}
}

@article{wang2025optimizing,
  title={Optimizing large language model training using fp4 quantization},
  author={Wang, Ruizhe and Gong, Yeyun and Liu, Xiao and Zhao, Guoshuai and Yang, Ziyue and Guo, Baining and Zha, Zhengjun and Cheng, Peng},
  journal={arXiv preprint arXiv:2501.17116},
  year={2025}
}

@article{mxfp,
  title={Microscaling data formats for deep learning},
  author={Rouhani, Bita Darvish and Zhao, Ritchie and More, Ankit and Hall, Mathew and Khodamoradi, Alireza and Deng, Summer and Choudhary, Dhruv and Cornea, Marius and Dellinger, Eric and Denolf, Kristof and others},
  journal={arXiv preprint arXiv:2310.10537},
  year={2023}
}

@article{hadamard,
  title={Outlier suppression: Pushing the limit of low-bit transformer language models},
  author={Wei, Xiuying and Zhang, Yunchen and Zhang, Xiangguo and Gong, Ruihao and Zhang, Shanghang and Zhang, Qi and Yu, Fengwei and Liu, Xianglong},
  journal={Advances in Neural Information Processing Systems},
  volume={35},
  pages={17402--17414},
  year={2022}
}

@misc{amd_mi355x_specs,
  title        = {AMD Instinct™ MI355X GPUs},
  author       = {{Advanced Micro Devices, Inc.}},
  howpublished = {\url{https://www.amd.com/en/products/accelerators/instinct/mi350/mi355x.html}} 
 
}

@article{chmiel2025fp4,
  title={Fp4 all the way: Fully quantized training of llms},
  author={Chmiel, Brian and Fishman, Maxim and Banner, Ron and Soudry, Daniel},
  journal={arXiv preprint arXiv:2505.19115},
  year={2025}
}

@inproceedings{liu2023llm,
  title={Llm-fp4: 4-bit floating-point quantized transformers},
  author={Liu, Shih-yang and Liu, Zechun and Huang, Xijie and Dong, Pingcheng and Cheng, Kwang-Ting},
  booktitle={Proceedings of the 2023 conference on empirical methods in natural language processing},
  pages={592--605},
  year={2023}
}

@article{zhou2025towards,
  title={Towards efficient pre-training: Exploring fp4 precision in large language models},
  author={Zhou, Jiecheng and Tang, Ding and Fu, Rong and Hu, Boni and Xu, Haoran and Wang, Yi and Pei, Zhilin and Su, Zhongling and Liu, Liang and Zhang, Xingcheng and others},
  journal={arXiv preprint arXiv:2502.11458},
  year={2025}
}

@article{abecassis2025pretraining,
  title={Pretraining large language models with nvfp4},
  author={Abecassis, Felix and Agrusa, Anjulie and Ahn, Dong and Alben, Jonah and Alborghetti, Stefania and Andersch, Michael and Arayandi, Sivakumar and Bjorlin, Alexis and Blakeman, Aaron and Briones, Evan and others},
  journal={arXiv preprint arXiv:2509.25149},
  year={2025}
}

@article{cim2026diagnosing,
  title={Diagnosing FP4 inference: a layer-wise and block-wise sensitivity analysis of NVFP4 and MXFP4},
  author={Cim, Musa and Topcu, Burak and Kandemir, Mahmut Taylan},
  journal={arXiv preprint arXiv:2603.08747},
  year={2026}
}
